\documentclass[journal,twoside,web]{ieeecolor}
\usepackage{generic}
\usepackage{cite}
\usepackage{amsmath,amssymb,amsfonts}
\usepackage{array}
\usepackage{booktabs}
\usepackage{graphicx}
\usepackage{algorithm,algorithmic}
\usepackage{hyperref}
\hypersetup{hidelinks=true}

\usepackage{multirow}
\usepackage{pifont}

\newcommand{\cmark}{\ding{51}}%
\newcommand{\xmark}{\ding{55}}%

\usepackage{textcomp}
\def\BibTeX{{\rm B\kern-.05em{\sc i\kern-.025em b}\kern-.08em
    T\kern-.1667em\lower.7ex\hbox{E}\kern-.125emX}}
\begin{document}
\title{Bidirectional Fusion Guided by Cardiac Patterns for Semi-Supervised ECG Segmentation}
\author{Jeonghwa Lim$^*$, Minje Park$^*$, and Sunghoon Joo$^\dagger$
\thanks{This research was supported by a grant of the Korea Health Technology R\&D Project through the Korea Health Industry Development Institute (KHIDI), funded by the Ministry of Health \& Welfare, Republic of Korea (grant number: RS-2023-00265466).}
\thanks{$^*$ J. Lim and M. Park contributed equally to this work and are with VUNO Inc., Seoul, South Korea (e-mail: \{jeonghwa.lim, minje.park\}@vuno.co).}
\thanks{$^\dagger$ Corresponding author. S. Joo is with VUNO Inc., Seoul, South Korea (e-mail: sunghoon.joo@vuno.co).}
}

\maketitle

\begin{abstract}
    Accurate delineation of electrocardiogram (ECG), the segmentation of meaningful waveform features, is crucial for cardiovascular diagnostics. However, the scarcity of annotated data poses a significant challenge for training deep learning models. Conventional semi-supervised semantic segmentation (SemiSeg) methods primarily focus on consistency from unlabeled data, underutilizing the information exchange possible between labeled and unlabeled sets. To address this, we introduce \textbf{CardioMix}, a framework built on a bidirectional CutMix strategy guided by cardiac patterns for ECG segmentation. This approach enriches the labeled set with realistic variations from unlabeled data while simultaneously applying stronger supervisory signals to the unlabeled set, as the cardiac pattern-guided mixing ensures all augmented samples remain physiologically meaningful. Our framework is designed as a plug-and-play module, demonstrating high compatibility with various SemiSeg algorithms.
    Extensive experiments on \textsf{SemiSegECG}, a public multi-dataset benchmark for ECG delineation, demonstrate that \textbf{CardioMix} consistently outperforms existing CutMix-based fusion strategies across diverse datasets and labeled ratios as a plug-and-play module compatible with various SemiSeg algorithms.
\end{abstract}


\section{Introduction}
\label{sec:introduction}
\IEEEPARstart{A}{n} electrocardiogram (ECG) is an essential diagnostic tool for monitoring cardiovascular diseases via tracking the electrical activity of the heart. To interpret ECGs and diagnose diseases, several key waveform features should be identified. ECG delineation (i.e., ECG segmentation) is the basic step that divides the ECG signal into the primary waveforms, P-wave, QRS complex, and T-wave, indicating the atrial depolarization, ventricular depolarization, and ventricular repolarization, respectively~\cite{gacek2011ecg}.

Applying deep learning to this task is a promising direction~\cite{jimenez2019u,liang2022ecg_segnet,joung2024deep}, but its progress is challenged by the difficulty of obtaining large-scale, expertly-annotated datasets. The high cost and time required for manual labeling result in a scarcity of labeled data. This situation, a small pool of labeled ECGs versus a large, accessible pool of unlabeled ECGs, makes semi-supervised semantic segmentation (SemiSeg) a particularly appropriate strategy~\cite{pelaez2023survey}.

The dominant paradigm in SemiSeg is self-training with consistency regularization, where a model generates pseudo-labels for unlabeled data and is trained to be consistent against perturbations. To improve the quality of unreliable pseudo-labels, recent state-of-the-art methods in computer vision employ mixing strategies~\cite{zhao2023augmentation,fang2023locating}. Specifically, they use CutMix~\cite{yun2019cutmix} to paste patches from labeled images onto unlabeled ones, thereby providing more reliable supervision. However, these methods are ill-suited for ECG signals.

\begin{figure*}[h]
    \centering
    \includegraphics[width=1.0\linewidth]{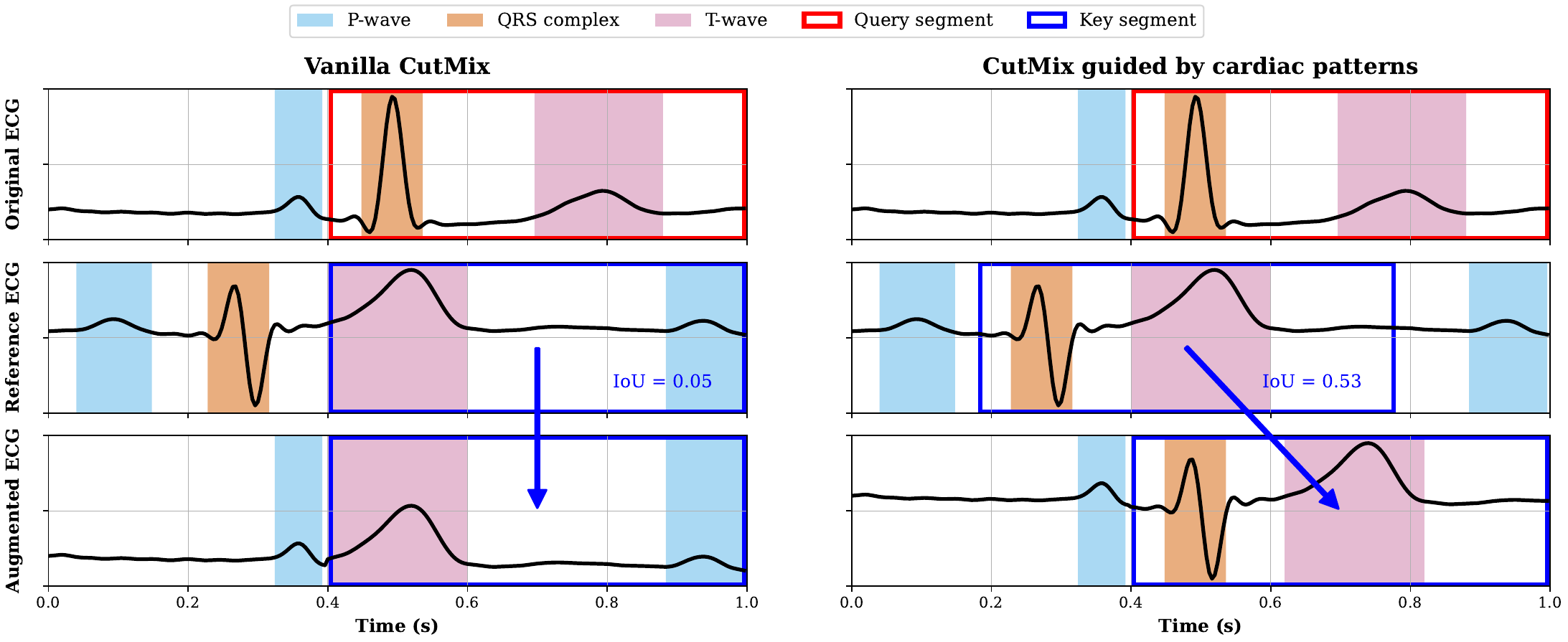}
    \caption{
        Cardiac pattern-guided CutMix for semi-supervised ECG segmentation. Vanilla CutMix (left) can create waveforms with inconsistent cardiac patterns by pasting random segments, resulting in an unnatural sequence (e.g., a P-wave followed by a T-wave). The proposed method, CutMix guided by cardiac patterns (right), prevents this by matching an original ECG segment (query) with a reference ECG segment (key) whose cardiac pattern is the most similar.
    }
    \label{fig:motivation}
\end{figure*}

Fig.~\ref{fig:motivation} illustrates the challenge of applying CutMix to ECG signals. Random segment selection can yield dissimilar matching (e.g., IoU=0.05), producing sequences like \texttt{P-T-P} that disrupt the natural \texttt{P-QRS-T} cardiac pattern. Furthermore, we identify that these unidirectional fusion methods are suboptimal, only augmenting unlabeled data and thus missing the opportunity to diversify the labeled set with variations from unlabeled signals.

To overcome these challenges, we propose \textbf{CardioMix}, a semi-supervised learning framework specifically designed for ECG delineation. \textbf{CardioMix} is built upon two core principles: (1) \textbf{Cardiac Pattern-Guided Search}: We guide segment selection using cardiac pattern similarity, measured via segmentation IoU. This ensures segments with similar waveform compositions are matched, preserving natural physiological structure. (2) \textbf{Bidirectional Fusion}: By enabling bidirectional information exchange between labeled and unlabeled data, the labeled set is enriched with diverse patterns from high-confidence unlabeled data, while unlabeled samples receive stronger supervisory signals from labeled segments, improving both supervised and unsupervised losses.

Our main contributions are as follows:
\begin{itemize}
\item We propose \textbf{CardioMix}, a new semi-supervised ECG delineation framework that can integrate with existing algorithms and address conventional limitations.
\item Through extensive experiments on public ECG datasets, we demonstrate that the proposed method outperforms existing approaches across various scenarios.
\item We present quantitative and qualitative analyses, showing that \textbf{CardioMix} maintains the highest structural consistency compared to baseline methods.
\end{itemize}

\section{Backgrounds}
\subsection{ECG Delineation}
Traditionally, ECG delineation has been addressed with heuristic, signal processing-based methods designed to locate fiducial points on the waveform~\cite{martinez2004wavelet}. While effective to a degree, these approaches often struggle with the noise and high variability inherent in clinical recordings, such as motion artifacts and baseline wandering~\cite{friesen1990comparison,elgendi2014revisiting,liu2018performance}.

In recent years, deep learning techniques have been applied to ECG delineation, demonstrating remarkable performance improvements by learning complex patterns directly from annotated signals~\cite{jimenez2019u,liang2022ecg_segnet,joung2024deep}. However, the vast majority of the deep learning models are fully supervised, relying on large-scale, high-quality labeled datasets. This dependency remains a critical bottleneck for the widespread deployment of deep learning in ECG delineation due to the costly and time-consuming nature of expert annotation.

\subsection{Semi-Supervised Semantic Segmentation}
The challenge of acquiring dense, point-level annotations makes SemiSeg a particularly well-suited paradigm for the medical domain, such as ECG delineation, where unlabeled data is far more abundant than labeled data~\cite{goldberger2000physiobank}. The dominant approaches in SemiSeg are using consistency regularization, where a model is trained to produce stable predictions under perturbations~\cite{pelaez2023survey}. This principle is central to foundational methods like \textbf{Mean Teacher} (\textbf{MT})~\cite{tarvainen2017mean}, which uses an exponential moving average (EMA) teacher to generate stable targets, and \textbf{FixMatch}~\cite{sohn2020fixmatch}, which leverages weak and strong augmentations. This concept has been extended in various ways, for instance by using dual networks to provide mutual supervision, as in \textbf{Cross Pseudo Supervision} (\textbf{CPS})~\cite{chen2021semi}, incorporating contrastive learning to improve feature representations at the region level, as in \textbf{Regional Contrast} (\textbf{ReCo})~\cite{liu2022bootstrapping}, or employing multi-stage training to progressively refine pseudo-labels, as in \textbf{Self-Training++} (\textbf{ST++})~\cite{yang2022st++}.

\subsection{Consistency Regularization with CutMix}
CutMix~\cite{yun2019cutmix} is widely adopted as a standard perturbation method for consistency regularization in SemiSeg, owing to its robust performance~\cite{french2020semi}. The conventional approach involves applying CutMix between samples within the unlabeled set, which serves as a strong and effective data augmentation strategy for consistency regularization.

More recent methods have evolved this idea to directly improve the quality of pseudo-labels by mixing information from the labeled set into the unlabeled set, which we call labeled-to-unlabeled fusion (L2U). For instance, \textbf{AugSeg} performs L2U, gating the operation with an adaptive confidence threshold, applying it only when the model's pseudo-label for the unlabeled sample is sufficiently reliable~\cite{zhao2023augmentation}. Similarly, \textbf{Uncertainty-aware Patch CutMix} (\textbf{UPC}) strategically identifies regions in unlabeled data with high prediction uncertainty and replaces them with random patches from a labeled image~\cite{fang2023locating}.

However, these methods leave two opportunities unexplored. First, they employ random segment selection without preserving physiological structure. Second, their unidirectional fusion only augments unlabeled data, missing opportunities to diversify the labeled set.

\section{Method}
\begin{figure}[t]
    \centering
    \includegraphics[width=0.65\columnwidth]{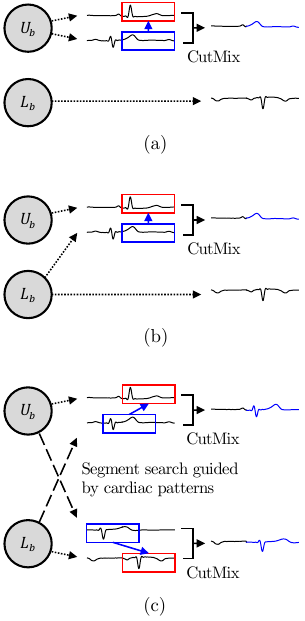}
    \caption{
        A taxonomy of CutMix-based fusion methods.
        The red box is a segment to be replaced (query), and the blue box is a segment to be fused (key) in the place of the query.
        $U_{b}$ and $L_b$ are unlabeled and labeled mini-batches, respectively.
        (a) Vanilla CutMix. (b) Labeled-to-unlabeled fusion. (c) \textbf{CardioMix}, a bidirectional fusion guided by cardiac patterns.
    }
    \label{fig2}
\end{figure}

In this section, we first review the underlying consistency regularization framework and then detail the proposed fusion mechanism. The \textbf{CardioMix}, as illustrated in Fig.~\ref{fig2}, consists of a cardiac pattern–guided segment search and a bidirectional fusion to address limitations of existing SemiSeg methods for ECG delineation.

\subsection{Problem Formulation}
We consider a labeled set $L=\{(x_{i}, y_{i})\}_{i=1}^{N_{L}}$ and an unlabeled set $U=\{u_{i}\}_{i=1}^{N_{U}}$ where each sample $x, u\in\mathbb{R}^{T}$ is a single-lead ECG signal of length T, and the label $y\in\{1,..., C\}^{T}$ denotes one of $C$ classes (e.g., background, P-wave, QRS complex, and T-wave). The proposed approach operates within a teacher-student framework, where the student model $f(\cdot;\theta)$ is trained on both labeled and unlabeled data. The teacher model, denoted as $f(\cdot;\theta^-)$, provides pseudo-label targets. The overall loss is a combination of a supervised loss $\mathcal{L}_s$ and a consistency loss $\mathcal{L}_c$: $\mathcal{L} = \mathcal{L}_{s} + \lambda\mathcal{L}_{c}$, where $\lambda$ is a weighting coefficient. The supervised loss is the standard cross-entropy loss $\mathcal{L}_{CE}$ on labeled mini-batch $L_b$: 
{
\begin{equation*}
    \mathcal{L}_{s} = \frac{1}{|L_{b}|}\sum_{(x,y)\in L_{b}}\mathcal{L}_{CE}(f(x;\theta),y)
\end{equation*}
}
The consistency loss enforces stable predictions on an augmented unlabeled sample $\tilde{u}$ using the corresponding pseudo-label $\tilde{p}$:
{
\begin{equation*}
    \mathcal{L}_{c} = \frac{1}{|U_{b}|}\sum_{u\in U_{b}}\mathcal{L}_{U}(f(\tilde{u};\theta), \tilde{p})
\end{equation*}
}
where $U_{b}$ is an unlabeled mini-batch, and $\mathcal{L}_{U}$ is a task-specific loss function defined by each SemiSeg method. $\tilde{p}$ is derived from the initial pseudo-label $\hat{p}_{i}=h(f(u_{i};\theta^-))$, where $h(\cdot)$ denotes a pseudo-labeling function (e.g., \texttt{argmax}).

\textbf{AugSeg} and \textbf{UPC} adopt labeled-to-unlabeled fusion (L2U), which generate the augmented pair $(\tilde{u}_{i},\tilde{p}_{i})$ by replacing a segment in an unlabeled sample $u_{i}$ with a segment from a labeled sample $x_{j}$. A segment is defined by an interval of time indices $[s,e]$, representing the set $\{s,s+1,...,e\}$. For a random segment $[s,e]$, the components of the augmented sequences at each time step $t\in\{1,...,T\}$ are defined as:
{
\begin{equation*}
    (\tilde{u}_{i,t}, \tilde{p}_{i,t})=\begin{cases}
        (x_{j,t}, y_{j,t}), & s\leq t\leq e \\
        (u_{i,t}, \hat{p}_{i,t}), & \text{otherwise}
    \end{cases}
\end{equation*}
}
The critical flaw in this approach, however, emerges when applied to structured signals like ECGs. A random segment selection can easily disrupt the natural cardiac cycle, necessitating a more informed selection strategy.

\subsection{Cardiac Pattern-Guided Search}
To ensure the physiological consistency, we employ a cardiac pattern-guided search strategy. Rather than random segment selection, the proposed method identifies segments with similar cardiac waveform patterns based on their composition in the segmentation space.

We measure pattern similarity using segmentation-space representations, as they provide a more stable and class-aware measure than raw signal features. Specifically, we use class-averaged intersection over union (IoU) as the metric for pattern similarity, which captures the overlap in waveform composition (P-wave, QRS complex, T-wave) between two segments. Assume two label segments of length $W$: $y_a=(y_{a,t})_{t=1}^{W}$ and $y_b=(y_{b,t})_{t=1}^{W}$. We define the pattern similarity $\mathrm{Sim}(y_a,y_b)$ as:
{
\begin{gather*}
    \mathrm{Sim}(y_{a},y_{b})=\frac{1}{C}\sum_{c=1}^{C}\mathrm{IoU}_{c}(y_{a},y_{b}),\\
    \mathrm{IoU}_{c}(y_{a},y_{b})=\frac{\sum_{t=1}^{W}\mathbb{I}(y_{a,t}=c\land y_{b,t}=c)}{\sum_{t=1}^{W}\mathbb{I}(y_{a,t}=c\lor y_{b,t}=c)}
\end{gather*}
}
where $\mathbb{I}(\cdot)$ is the indicator function. The process is then as follows: for each unlabeled sample $u_i$ in a mini-batch, we randomly select a query segment $[s_{q},e_{q}]$ to be replaced. We then search for a key segment across all labeled samples $(x_j,y_j)$ in the mini-batch $L_{b}$ that best matches the query's pseudo-label segment, $(\hat{p}_{i,t})_{t=s_{q}}^{e_{q}}$. 

Let $\mathcal{S}_{\mathrm{L}}=\{(j,[s_k,e_k])\,|\,(x_j,y_j)\in L_b, [s_k,e_k]\in\mathcal{W}\}$ be the search space of all possible key segments from the labeled mini-batch, where $\mathcal{W}$ is the set of all possible segments of a fixed length $W$ generated with a stride $S_w$. The optimal labeled sample and key segment are thus found by maximizing the similarity over this set:

\begin{equation}
\begin{aligned}
    &(j^*,[s_k^*,e_k^*]) \\
    &\quad = \mathop{\mathrm{arg\,max}}_{(j,[s_k,e_k])\in\mathcal{S}_{L}}\mathrm{Sim}\left((\hat{p}_{i,t})_{t=s_{q}}^{e_{q}}, (y_{j,t})_{t=s_{k}}^{e_{k}}\right)
\end{aligned}
\label{eq:search_usl}
\end{equation}

The final augmentation is performed by pasting the optimal segment from the labeled sample $x_{j^*}$ into the query location. For each time step $t\in\{1,...,T\}$, the components of the final augmented sequences $(\tilde{u}_{i,t},\tilde{p}_{i,t})$ are defined as:
\begin{equation}
\begin{aligned}
    &(\tilde{u}_{i,t}, \tilde{p}_{i,t}) \\
    &\quad = \begin{cases}
        (x_{j^*,t-s_q+s_k^*}, y_{j^*,t-s_q+s_k^*}), & s_q\leq t\leq e_q \\
        (u_{i,t}, \hat{p}_{i,t}), & \text{otherwise}
    \end{cases}
\end{aligned}
\label{eq:replace_usl}
\end{equation}
This approach naturally yields segments with valid cardiac patterns, preserving the physiological structure of the cardiac cycle.

\subsection{Bidirectional Fusion with Confidence Gating}
The fusion process described thus far is unidirectional. To maximize information exchange, we introduce a reversed, unlabeled-to-labeled fusion (U2L) strategy. This acts as a sophisticated data augmentation for the labeled set. Meanwhile, pasting segments with potentially noisy pseudo-labels onto clean labeled data risks corrupting the ground truth. To mitigate this, we also introduce a confidence gating mechanism. The U2L is only performed if the teacher model is highly confident about its prediction for the selected unlabeled segment.

The U2L process mirrors the L2U. For each labeled sample $(x_i,y_i)$, we randomly select a query segment $[s_q,e_q]$ and find the key segment $[s_k^*,e_k^*]$ with the most similar cardiac pattern from unlabeled samples $u_{j^*}$ in the mini-batch $U_b$.

Let $\mathcal{S}_{\mathrm{U}}=\{(j,[s_k,e_k])\,|\,u_{j}\in U_b, [s_k,e_k]\in\mathcal{W}\}$ be the reverse search space. The optimal unlabeled sample and key segment are found:
\begin{equation}
\begin{aligned}
    &(j^*,[s_k^*,e_k^*]) \\
    &\quad = \mathop{\mathrm{arg\,max}}_{(j,[s_k,e_k])\in\mathcal{S}_{U}}\mathrm{Sim}\left((y_{i,t})_{t=s_{q}}^{e_{q}}, (\hat{p}_{j,t})_{t=s_{k}}^{e_{k}}\right)
\label{eq:search_lsu}
\end{aligned}
\end{equation}
We then compute the confidence of the teacher's prediction for this optimal segment. Let $\left(\sigma_{c}(u_{j^*,t};\theta^-)\right)_{c=1}^{C}$ be the softmax probability output of the teacher model for the time step $t$. The segment confidence is the average maximum probability within the segment:
\begin{equation}
    \mathrm{Conf}(u_{j^*},[s_k^*,e_k^*])=\frac{1}{W}\sum_{t=s_k^*}^{e_k^*}\max_{c}\left(\sigma_{c}(u_{j^*,t};\theta^-)\right)_{c=1}^{C}
\label{eq:conf}
\end{equation}
The U2L is gated by a confidence threshold $\tau$. If the confidence exceeds $\tau$, we generate an augmented labeled pair $(\tilde{x}_{i},\tilde{y}_{i})$. For each time step $t$, this augmentation is defined as:
\begin{equation}
\begin{aligned}
    &(\tilde{x}_{i,t}, \tilde{y}_{i,t}) \\
    & \quad = \begin{cases}
        (u_{j^*,t-s_q+s_k^*}, \hat{p}_{j^*,t-s_q+s_k^*}), & s_q\leq t\leq e_q \\
        (x_{i,t}, y_{i,t}), & \text{otherwise}
    \end{cases}
\end{aligned}
\label{eq:replace_lsu}
\end{equation}

If the confidence is below the threshold, the original labeled pair is used, i.e., $(\tilde{x}_{i},\tilde{y}_{i})=(x_{i},y_{i})$. The supervised loss is then computed over the batch of potentially augmented labeled pairs. This bidirectional approach allows for a more robust exchange of information, improving both consistency regularization and supervised training.

The complete \textbf{CardioMix} framework integrates all proposed components: cardiac pattern-guided search, bidirectional fusion, and confidence gating strategy. A pseudo-code describing the full procedure of \textbf{CardioMix} is presented in Algorithm~\ref{alg:cardiomix}.

\begin{algorithm}[t]
\caption{The \textbf{CardioMix} Framework}
\label{alg:cardiomix}
\small
\textbf{Input}: Labeled set $L$, unlabeled set $U$, student model $f(\cdot;\theta)$, teacher model $f(\cdot;\theta^-)$. \\
\textbf{Hyperparameter}: coefficient $\lambda$, confidence threshold $\tau$, window size $W$, stride $S_w$.\\
\textbf{Output}: Updated student $f(\cdot;\theta)$. \\
\begin{algorithmic}[1] 
\FOR{each training step}
\STATE Sample mini-batch $L_b=\{(x_{i},y_{i})\}_{i=1}^{|L_{b}|}$ from $L$.
\STATE Sample mini-batch $U_b=\{u_{i}\}_{i=1}^{|U_{b}|}$ from $U$.
\STATE Generate initial pseudo-labels $\hat{p}_{i}=h(f(u_{i};\theta^-))$ for all $u_{i}\in U_b$.
\STATE
\STATE // Labeled-to-unlabeled fusion
\FOR{each unlabeled sample $u_i\in U_b$}
\STATE Randomly select a query segment $[s_q,e_q]$.
\STATE Find $(j^*,[s_k^*,e_k^*])$ by Eq.~\ref{eq:search_usl}.
\STATE Construct augmented pair $(\tilde{u}_i,\tilde{p}_i)$ by Eq.~\ref{eq:replace_usl}.
\ENDFOR
\STATE
\STATE // Unlabeled-to-labeled fusion
\FOR{each labeled sample $(x_i,y_i)\in L_b$}
\STATE Randomly select a query segment $[s_q,e_q]$.
\STATE Find $(j^*,[s_k^*,e_k^*])$ by Eq.~\ref{eq:search_lsu}.
\STATE Calculate confidence $\mathrm{Conf}$ by Eq.~\ref{eq:conf}.
\IF {$\mathrm{Conf}>\tau$}
\STATE Construct augmented pair $(\tilde{x}_i,\tilde{y}_i)$ by Eq.~\ref{eq:replace_lsu}.
\ELSE
\STATE $(\tilde{x}_i,\tilde{y}_i)\leftarrow(x_i,y_i)$.
\ENDIF
\ENDFOR
\STATE
\STATE $\mathcal{L}_s\leftarrow \frac{1}{|L_b|}\sum_{i=1}^{|L_b|}\mathcal{L}_{CE}(f(\tilde{x}_{i};\theta),\tilde{y}_{i})$.
\STATE $\mathcal{L}_c\leftarrow \frac{1}{|U_b|}\sum_{i=1}^{|U_b|}\mathcal{L}_{U}(f(\tilde{u}_{i};\theta),\tilde{p}_{i})$.
\STATE $\mathcal{L}\leftarrow \mathcal{L}_s+\lambda\mathcal{L}_c$.
\STATE Update student parameters $\theta$ by backpropagating $\mathcal{L}$.
\STATE Update teacher parameters $\theta^-$ (e.g., via EMA).
\ENDFOR
\STATE \textbf{return} $f(\cdot;\theta)$.
\end{algorithmic}
\end{algorithm}

\begin{table*}[t]
    \small
    \centering
    \caption{
        Characteristics of the ECG databases. 12-lead ECGs comprise limb (I, II, III, aVR, aVL, aVF) and precordial (V1–V6) leads; and the 2-lead ECGs consist of any two selected leads (e.g., MLII and V1).
    }
    \begin{tabular}{l|c|c|c|c|c|c|ccc}
    \toprule
    \multirow{2}{*}{\textbf{Source}}&  \multirow{2}{*}{\textbf{\# Subjects}}&\multirow{2}{*}{\textbf{\# ECGs}} & \multirow{2}{*}{\textbf{Duration (labeled)}}& \multirow{2}{*}{\textbf{Sample rate}}&\multirow{2}{*}{\textbf{Lead type}}& \multicolumn{4}{c}{\textbf{\# Samples}}\\
    &   && & &  &  All&Train& Validation&Test\\
    \midrule
    \textit{LUDB} &  200&200  & 10 s& 500 Hz & 12-lead &  2,369&1,427& 468&474\\
    \textit{QTDB} &  105&105  & 5.9--253.6 s& 250 Hz & 2-lead &  718&422& 148&148\\
    \textit{ISP} &  499&499  & 10 s& 1000 Hz & 12-lead &  5,988&3,792& 1,272&924\\
    \textit{Zhejiang} &  334&334  & 1.3--7.1 s& 2000 Hz & 12-lead &  4,008&2,400& 804&804\\
    \bottomrule
    \end{tabular}
    \label{tab:ecg_dataset}
\end{table*}

\section{Experiments}
\subsection{Baseline Algorithms}
The experiments are designed to demonstrate that \textbf{CardioMix} acts as a superior, plug-and-play data fusion module. To this end, we first establish a set of diverse base algorithms: \textbf{MT}~\cite{tarvainen2017mean}, \textbf{FixMatch}~\cite{sohn2020fixmatch}, \textbf{CPS}~\cite{chen2021semi}, \textbf{ReCo}~\cite{liu2022bootstrapping}, and \textbf{ST++}~\cite{yang2022st++}. On top of each of these base algorithms, we compare the performance of several CutMix-based fusion strategies:
\begin{itemize}
    \item \textbf{CutMix}~\cite{french2020semi}, which mixes two unlabeled samples.
    \item \textbf{AugSeg}~\cite{zhao2023augmentation}, which gates L2U randomly based on the confidence of the pseudo-label.
    \item \textbf{UPC}~\cite{fang2023locating}, which performs L2U on regions with the highest uncertainty.
    \item \textbf{CardioMix}, the proposed bidirectional fusion strategy guided by cardiac patterns.
\end{itemize}
By applying each fusion strategy to every base algorithm, we can fairly evaluate the added benefit of the proposed approach across a wide range of modern SemiSeg frameworks. For clarity, we first compare fusion strategies using \textbf{MT} as a representative base algorithm (Section~\ref{sec:sota}), followed by a full plug-and-play evaluation across all base algorithms (Section~\ref{sec:plug}).

\subsection{Databases and Preprocessing}
We evaluate \textbf{CardioMix} on four publicly available ECG databases: \textit{LUDB}~\cite{kalyakulina2020ludb}, \textit{QTDB}~\cite{laguna1997database}, \textit{ISP}~\cite{avetisyan2024ispdataset}, and \textit{Zhejiang}~\cite{zheng202012}. While \textit{LUDB} and \textit{QTDB} are well-established benchmarks in ECG delineation, \textit{ISP} and \textit{Zhejiang} are more recent resources that offer rich segmentation annotations. A summary of the databases is provided in Table~\ref{tab:ecg_dataset}.
\subsubsection{\textit{LUDB}} This database has served as a primary benchmark in ECG delineation studies due to its high-quality annotations. A key advantage of \textit{LUDB} is its provision of lead-specific delineation labels. Unlike other databases that often share a single annotation across all leads for the same sample, \textit{LUDB} provides distinct annotations for each of the 12 leads. 

\subsubsection{\textit{QTDB}} This database provides ECG recordings with specific 2-lead. We utilize the revised annotations from a prior work~\cite{jimenez2024generalising}, which were introduced to resolve critical issues in the original labels, such as missing beats and incomplete onset-offset markings.

\subsubsection{\textit{ISP}} The \textit{ISP} database was introduced to overcome common limitations in existing delineation datasets, including small scale, low noise levels, and limited rhythm diversity. It provides 12-lead ECGs with detailed annotations, supporting the development of more robust and generalizable models. 

\subsubsection{\textit{Zhejiang}} Originally released for idiopathic ventricular arrhythmia classification, the \textit{Zhejiang} database contains 12-lead ECGs. Its use in segmentation tasks for this work was enabled by the new delineation ground truth introduced in a prior study~\cite{jimenez2024generalising}.

All ECG signals are fixed to a 10-second duration by cropping or zero-padding if needed. They are resampled to 250 Hz and applied a band-pass filter (0.67–40 Hz) to remove baseline wander and high-frequency noise. Finally, we also apply Z-score normalization to each signal before feeding it into the models. 

\subsection{Experimental Protocols}
To ensure a fair and reproducible comparison, we adopt the standardized benchmark setting, \textsf{SemiSegECG}, from a previous open benchmark study~\cite{park2025semisegecg}. We adhere to its configurations, which include using the predefined subject-wise data splits (train, validation, and test), the provided indices for various labeled data ratios (1/16, 1/8, 1/4, and 1/2), and treating each ECG lead as an independent sample.

In terms of performance assessment, we employ the mean Intersection over Union (mIoU) as the primary metric to measure the overall segmentation accuracy. We also assess clinical utility through the mean absolute errors (MAEs) of key diagnostic intervals (e.g., PR interval, QRS duration, and QT interval). The reported performance corresponds to the checkpoints that achieved the highest mIoU on the validation set. All experiments are conducted three times with different random seeds, and we report the averages and standard deviations of metrics from these runs.

To comprehensively evaluate the effectiveness and robustness of \textbf{CardioMix}, we conduct experiments under two distinct scenarios: an in-domain and a cross-domain.
\subsubsection{In-domain}
This setting evaluates the inherent effectiveness of the method by training and testing on data from the same source, free from the confounding effects of domain shift. In this protocol, the evaluation is conducted across all four datasets and labeled ratios, serving as the primary benchmark for assessing the performance under varying degrees of label scarcity.

\subsubsection{Cross-domain}
This setting assesses the ability of the method to generalize across domains by leveraging unlabeled data from a different domain.
In this protocol, \textit{QTDB} provides the labeled set for training, validation, and testing, while the entire \textit{ISP} dataset, approximately 14 times larger than the \textit{QTDB} training set, serves as a large-scale unlabeled source, simulating a practical scenario where a small, expert-annotated dataset is augmented with a large pool of unlabeled signals from a different acquisition environment.

\subsection{Implementation Details}
We follow the \textsf{SemiSegECG} benchmark study for the overall training recipes, including training schedule, augmentations, and SemiSeg-specific hyperparameters. All experiments are conducted using PyTorch 1.11 and Python 3.9 with four NVIDIA RTX 4090 GPUs.

\subsubsection{Model Architecture} 
We choose ViT as the encoder, motivated by its reliable performance in ECG analysis~\cite{li2021bat,na2024guiding} and its effectiveness in semi-supervised ECG segmentation~\cite{park2025semisegecg}. Specifically, we adopt ViT-Tiny~\cite{touvron2021training} as the encoder and a two-layer FCN~\cite{long2015fully} as the decoder, with a 128-dimensional hidden layer and a dropout rate of $p = 0.1$~\cite{srivastava2014dropout}. 

\subsubsection{Training schedule} All models are trained for 100 epochs with a batch size of 16, using the AdamW optimizer~\cite{loshchilov2018decoupled} with a weight decay of 0.05. The learning rate is warmed up from 0 to 0.001 over the first 10 epochs and then decayed to 0.0001 using cosine annealing~\cite{loshchilov2017sgdr}.

\subsubsection{Augmentations} Random resized cropping is adopted as the weak augmentation. For strong augmentation, we employ a RandAugment policy~\cite{cubuk2020randaugment}, where any three of the following four transformations are randomly selected and applied to each sample: powerline noise, sine-wave noise, amplitude scaling, and white noise.

\subsubsection{SemiSeg-specific hyperparameters} For EMA-based models (\textbf{MT}, \textbf{ReCo}, \textbf{ST++)}, the decay rate is set to 0.99. The confidence threshold is 0.8 for \textbf{FixMatch}, and 0.65 (easy) and 0.8 (hard) for \textbf{ReCo}. The projection head used in \textbf{ReCo} consists of two convolutional layers with 128 channels each.

\subsubsection{CutMix hyperparameters} The segment window size $W$ is randomly sampled for each mini-batch in $[250,1250]$ (1s--5s), and the stride $S_w$ is $W/2$. For \textbf{CardioMix}, the confidence gating threshold $\tau$ is set to 0.8 based on the grid search, with a detailed sensitivity analysis provided in Section~\ref{sec:ablation}.

\begin{table}[t]
\centering
\caption{
    Benchmarking results (mIoU, \%) of CutMix-based fusion strategies (\textbf{MT} as base algorithm).
}
\begin{tabular}{l|cccc}
\toprule
\multirow{2}{*}{\textbf{Method}} & \multicolumn{4}{c}{\textbf{Labeled ratio}} \\
& 1/16 & 1/8 & 1/4 & 1/2 \\
\midrule
\multicolumn{5}{c}{\textit{LUDB}\hspace{0.15cm}(\textit{N} = 1,427)} \\
\midrule
 \textbf{MT}& 73.4±0.2& 76.7±0.1& 78.7±0.0&80.4±0.2\\
+ \textbf{CutMix}& 73.8±0.5& 77.0±0.2& 79.0±0.2& 81.0±0.2
\\
+ \textbf{AugSeg}& 73.8±0.3& 77.0±0.1& 79.5±0.3& 80.9±0.2
\\
+ \textbf{UPC}& 74.9±0.2& 77.4±0.3& 79.7±0.2& 81.7±0.3\\
+ \textbf{CardioMix}& \textbf{78.4±0.1}& \textbf{80.2±0.3}& \textbf{82.2±0.1}& \textbf{83.5±0.1}\\
\midrule
\multicolumn{5}{c}{\textit{QTDB}\hspace{0.15cm}(\textit{N} = 422)} \\
\midrule
 \textbf{MT}& 55.1±0.2& 56.9±1.0& 64.9±0.2&69.4±0.2\\
+ \textbf{CutMix}& 55.1±0.4& 57.8±1.3& 66.1±0.9& 69.2±0.4
\\
+ \textbf{AugSeg}& 55.9±1.0& 57.3±1.0& 66.5±0.4& 69.9±0.9
\\
+ \textbf{UPC}& 56.9±0.8& 58.7±0.9& 66.0±0.7& 70.0±0.4\\
+ \textbf{CardioMix}& \textbf{60.2±0.6}& \textbf{62.8±0.4}& \textbf{68.4±0.5}& \textbf{71.3±0.3}\\
\midrule
\multicolumn{5}{c}{\textit{ISP}\hspace{0.15cm}(\textit{N} = 3,792)} \\
\midrule
 \textbf{MT}& 74.6±0.1& 75.7±0.3& 78.1±0.3&78.8±0.3\\
+ \textbf{CutMix}& 72.9±0.4& 75.0±0.3& 78.4±0.4& 80.1±0.2
\\
+ \textbf{AugSeg}& 73.3±0.6& 75.2±0.7& 78.3±0.4& 80.1±0.5
\\
+ \textbf{UPC}& 73.8±0.1& 76.2±0.3& 79.6±0.7& 81.6±0.4\\
+ \textbf{CardioMix}& \textbf{75.0±0.7}& \textbf{78.9±0.3}& \textbf{81.4±0.2}& \textbf{82.5±0.1}\\
\midrule
\multicolumn{5}{c}{\textit{Zhejiang}\hspace{0.15cm}(\textit{N} = 2,400)} \\
\midrule
 \textbf{MT}& 78.8±0.8& 81.1±0.3& 81.8±0.4&83.5±0.1\\
+ \textbf{CutMix}& 78.9±0.6& 81.0±0.0& 82.2±0.1& 83.6±0.3
\\
+ \textbf{AugSeg}& 
79.4±0.5& 81.5±0.3& 82.6±0.4& 83.6±0.2
\\
+ \textbf{UPC}& 79.8±0.1& \textbf{82.1±0.2}& 82.9±0.3& \textbf{83.8±0.3}\\
+ \textbf{CardioMix}& 
\textbf{81.1±0.1}& 82.0±0.0& \textbf{82.9±0.1}& 83.7±0.0\\
\bottomrule
\end{tabular}
\label{tab:main-sota}
\end{table}

\begin{table}[t]
\centering
\caption{
    Benchmarking results (averaged interval MAE, ms) of CutMix-based fusion strategies (\textbf{MT} as base algorithm).
}
\begin{tabular}{l|cccc}
\toprule
\multirow{2}{*}{\textbf{Method}} & \multicolumn{4}{c}{\textbf{Labeled ratio}} \\
& 1/16 & 1/8 & 1/4 & 1/2 \\
\midrule
\multicolumn{5}{c}{\textit{LUDB}\hspace{0.15cm}(\textit{N} = 1,427)} \\
\midrule
 \textbf{MT}& 24.9±0.9& 18.2±0.2& 17.1±0.9&14.9±0.5
\\
+ \textbf{CutMix}& 24.1±3.5& 19.2±0.8& 16.8±1.3& 14.5±0.4
\\
+ \textbf{AugSeg}& 21.8±1.1& 17.8±1.4& 16.5±0.3& 14.2±0.4
\\
+ \textbf{UPC}& 20.9±1.9& 17.1±0.3& 16.7±0.8& 13.4±0.3
\\
+ \textbf{CardioMix}& \textbf{18.0±0.4}& \textbf{15.3±0.4}& \textbf{14.5±0.1}& \textbf{12.8±0.3}
\\
\midrule
\multicolumn{5}{c}{\textit{QTDB}\hspace{0.15cm}(\textit{N} = 422)} \\
\midrule
 \textbf{MT}& 67.6±5.0& 69.1±17.3& 61.3±9.4&45.5±3.2
\\
+ \textbf{CutMix}
& 60.9±2.7& 61.7±5.2& 56.1±10.3& 43.5±3.5
\\
+ \textbf{AugSeg}
& 62.8±4.5& 65.4±9.2& 46.4±4.2& 41.4±3.9
\\
+ \textbf{UPC}
& 48.1±3.4& 58.3±5.4& \textbf{43.5±1.0}& \textbf{40.7±2.0}
\\
+ \textbf{CardioMix}& \textbf{45.1±1.2}& \textbf{45.2±4.6}& 44.1±1.7& 41.0±4.0
\\
\midrule
\multicolumn{5}{c}{\textit{ISP}\hspace{0.15cm}(\textit{N} = 3,792)} \\
\midrule
 \textbf{MT}& \textbf{25.3±1.7}& 22.7±0.5& 21.4±1.5&20.5±1.0
\\
+ \textbf{CutMix}
& 26.6±0.9& 24.3±1.8& 20.4±0.5& 19.2±0.7
\\
+ \textbf{AugSeg}
& 26.2±0.5& 24.2±1.9& 20.5±1.1& 18.7±0.6
\\
+ \textbf{UPC}
& 25.5±0.5& 22.6±0.4& 20.2±1.4& 17.9±0.5
\\
+ \textbf{CardioMix}& 25.7±1.8& \textbf{21.9±1.1}& \textbf{19.0±0.5}& \textbf{16.5±0.4}
\\
\midrule
\multicolumn{5}{c}{\textit{Zhejiang}\hspace{0.15cm}(\textit{N} = 2,400)} \\
\midrule
 \textbf{MT}& 19.5±13.1& 18.1±12.1& 17.0±11.3&\textbf{15.3±10.1}
\\
+ \textbf{CutMix}
& 18.4±12.3& 18.1±12.1& 16.7±11.1& 15.8±10.4
\\
+ \textbf{AugSeg}
& 
18.6±12.4& 17.8±11.9& 16.3±10.8& 15.6±10.3
\\
+ \textbf{UPC}
& 19.8±13.3& 17.6±11.8& \textbf{16.2±10.7}& 15.6±10.3
\\
+ \textbf{CardioMix}& 
\textbf{17.7±11.8}& \textbf{17.4±11.6}& 16.8±11.1& 16.3±10.8
\\
\bottomrule
\end{tabular}
\label{tab:main-sota(MAEs)}
\end{table}

\section{Results}

\subsection{Comparison with CutMix-Based Fusion Strategies}
\label{sec:sota}
We first evaluate the effectiveness of \textbf{CardioMix} against other CutMix-based fusion strategies across multiple datasets, adopting \textbf{MT} as the base algorithm. We choose \textbf{MT} based on its superior performance among semi-supervised base algorithms reported in the \textsf{SemiSegECG} benchmark~\cite{park2025semisegecg}, allowing us to focus the comparison on the fusion strategies themselves.

Table~\ref{tab:main-sota} summarizes the different fusion strategies when applied to \textbf{MT}. The results show that \textbf{CardioMix} surpasses the other baselines in most settings. While simpler strategies like \textbf{CutMix} and more advanced unidirectional methods like \textbf{AugSeg} and \textbf{UPC} yield modest gains, \textbf{CardioMix} exhibits a clear advantage across nearly all datasets and labeled ratios. For instance, under the highly data-scarce condition of a 1/16 labeled ratio on the \textit{LUDB} dataset, \textbf{CardioMix} achieves an mIoU of 78.4\%, significantly outperforming standard \textbf{CutMix} (73.8\%), \textbf{AugSeg} (73.8\%), and \textbf{UPC} (74.9\%). The consistent performance gains across diverse scenarios support our claim that the bidirectional fusion guided by cardiac patterns can maximize the reliability and the utility of available data under label-scarce conditions. On the \textit{Zhejiang} dataset, \textbf{UPC} shows slight advantages, likely due to sparse labeling and extensive zero-padding, which reduces the effectiveness of cardiac pattern-guided fusion and leads to comparable performances across methods. 

Table~\ref{tab:main-sota(MAEs)} further assesses the clinical utility of the segmentation by reporting the MAE (ms) averaged across the PR interval, QRS duration, and QT interval. \textbf{CardioMix} demonstrates superiority over other CutMix-based fusion methods across most datasets and labeled ratios, confirming that its benefits extend beyond mIoU to clinically relevant performance. Notably, on the \textit{LUDB} dataset, \textbf{CardioMix} achieves superior MAE to vanilla \textbf{MT} (i.e., \textbf{MT} trained without any CutMix-based fusion) while using only half the labeled data. For instance, \textbf{MT} + \textbf{CardioMix} trained on 1/8 labeled data outperforms vanilla \textbf{MT} trained on 1/4 labeled data, highlighting the remarkable data efficiency of \textbf{CardioMix}.

\begin{table}[t]
\centering
\caption{
    \textit{LUDB} benchmarking results (mIoU, \%) of CutMix-based fusion strategies across all base algorithms.
}
\begin{tabular}{lcccc}
\toprule
\multirow{3}{*}{\textbf{Method}} & \multicolumn{4}{c}{\textbf{Labeled ratio}} \\
& 1/16 & 1/8 & 1/4 & 1/2 \\
& (N=84) & (N=180) & (N=360) & (N=711) \\
\midrule
\textbf{Scratch}& 65.6±0.4& 71.5±0.4& 75.8±0.3& 78.7±0.3
\\
\midrule
\textbf{MT} & 73.4±0.2& 76.7±0.1& 78.7±0.0& 80.4±0.2
\\
+ \textbf{CutMix}& 73.8±0.5& 77.0±0.2& 79.0±0.2& 81.0±0.2
\\
+ \textbf{AugSeg}& 73.8±0.3& 77.0±0.1& 79.5±0.3& 80.9±0.2
\\
+ \textbf{UPC}& 74.9±0.2& 77.4±0.3& 79.7±0.2& 81.7±0.3
\\
+ \textbf{CardioMix}& \textbf{78.4±0.1}& \textbf{80.2±0.3}& \textbf{82.2±0.1}& \textbf{83.5±0.1}\\
\midrule
\textbf{FixMatch}& 71.6±0.6& 76.1±0.5& 78.2±0.2& 80.1±0.1
\\
+ \textbf{CutMix}& 73.4±0.4& 77.0±0.3& 79.0±0.3& 80.6±0.3
\\
+ \textbf{AugSeg}& 73.1±0.3& 76.7±0.1& 78.9±0.4& 80.5±0.1
\\
+ \textbf{UPC}& 73.6±0.1& 77.3±0.1& 79.4±0.2& 80.9±0.1
\\
+ \textbf{CardioMix}& \textbf{76.8±0.2}& \textbf{79.3±0.1}& \textbf{80.7±0.2}& \textbf{82.1±0.1}
\\
\midrule
\textbf{CPS}& 70.5±0.4& 75.0±0.4& 78.2±0.2& 80.1±0.1
\\
+ \textbf{CutMix}& 74.1±0.3& 77.9±0.1& \textbf{79.7±0.2}& 81.0±0.2
\\
+ \textbf{AugSeg}& 74.4±0.2& 78.2±0.2& 79.6±0.2& 81.1±0.2
\\
+ \textbf{UPC}& 73.9±0.5& 78.1±0.1& \textbf{79.7±0.2}& \textbf{81.3±0.0}
\\
+ \textbf{CardioMix}& \textbf{75.0±0.2}& \textbf{78.3±0.4}& 79.1±0.1& 80.9±0.2
\\
\midrule
\textbf{ReCo} & 71.7±0.4& 74.0±0.3& 74.8±0.1& 75.1±0.2
\\
+ \textbf{CutMix}& 73.3±0.0& 75.3±0.2& 75.9±0.2& 76.2±0.0
\\
+ \textbf{AugSeg}& 73.3±0.3& 75.5±0.1& 76.1±0.2& 76.4±0.2
\\
+ \textbf{UPC}& 74.3±0.2& \textbf{76.8±0.2}& \textbf{77.5±0.2}& \textbf{77.9±0.1}
\\
+ \textbf{CardioMix}& \textbf{75.3±0.3}& 76.7±0.2& 76.8±0.2& 77.3±0.1
\\
\midrule
\textbf{ST++} &71.6±0.7& 76.2±0.2& 78.2±0.2& 80.1±0.1
\\
+ \textbf{CutMix}& 72.3±0.3& 76.8±0.4& 79.0±0.0 & 80.8±0.2
\\
+ \textbf{AugSeg}& 72.9±0.4& 77.0±0.1& 79.1±0.1& 80.8±0.1
\\
+ \textbf{UPC}& 73.4±0.2& 77.4±0.2& 79.4±0.0 & \textbf{81.2±0.1}\\
+ \textbf{CardioMix}& \textbf{73.8±0.1}& \textbf{78.1±0.1}& \textbf{79.5±0.1}& 80.8±0.0\\
\bottomrule
\end{tabular}

\label{tab:bench-ludb}
\end{table}

\subsection{Plug-and-Play Compatibility}
\label{sec:plug}
A key advantage of \textbf{CardioMix} is its compatibility as a plug-and-play module with various SemiSeg algorithms.
To demonstrate this, we integrate \textbf{CardioMix} into all five base algorithms, comparing against other CutMix-based fusion strategies. We report results on the \textit{LUDB} dataset as the primary benchmark, given its high-quality annotations, with results on the remaining datasets provided in Appendix~\ref{appendix-ext_rst}.

As shown in Table~\ref{tab:bench-ludb}, \textbf{CardioMix} outperforms other CutMix-based fusion strategies across most base algorithms and labeled ratios. Notably, \textbf{CardioMix} demonstrates remarkable data efficiency. With \textbf{FixMatch}, \textbf{CardioMix} with a 1/8 labeled ratio achieves an mIoU of 79.3\%, surpassing the base \textbf{FixMatch} model trained with twice the amount of labeled data (1/4 ratio, 78.2\%). This suggests that \textbf{CardioMix} can be more impactful than simply acquiring more labeled data. Consistent trends are observed across the remaining datasets.

Further, to simulate a realistic clinical scenario with data from different acquisition environments, we perform a cross-domain experiment, using \textit{QTDB} and \textit{ISP} as labeled and unlabeled sets, respectively. As shown in Table~\ref{tab:cross_domain}, applying \textbf{CardioMix} in this challenging setting yields significant performance gains across all base algorithms and demonstrates superior robustness even compared to \textbf{UPC}. This validates the generalizability of \textbf{CardioMix} even in heterogeneous data conditions, highlighting its real-world applicability.

\begin{table}[t]
\centering
\caption{
    Cross-domain benchmarking results (mIoU, \%) using \textit{QTDB} and \textit{ISP} as labeled and unlabeled sets. Parentheses indicate improvement over the base algorithm (w/o CutMix).
}
\begin{tabular}{lccc}
\toprule
\textbf{Method} & \textbf{w/o CutMix} & \textbf{w/ UPC} & \textbf{w/ CardioMix} \\
\midrule
\textbf{MT}          & 70.8& 73.0 (+2.2)& 75.7 (+4.9)
\\
\textbf{Fixmatch}    & 70.7& 73.0 (+2.3)& 75.8 (+5.1)
\\
\textbf{CPS}         & 71.2& 73.0 (+1.8)& 74.6 (+3.4)
\\
\textbf{ReCo}        & 70.9& 72.3 (+1.4)& 73.7 (+2.8)
\\
\textbf{ST++}        & 70.2& 72.3 (+2.1)& 73.5 (+3.3)
\\
\midrule
\textbf{Average}     & 70.8& 72.7 (+2.0)& \textbf{74.7 (+3.9)}\\
\bottomrule
\end{tabular}
\label{tab:cross_domain}
\end{table}

\begin{table}[t]
\centering
\caption{
    Ablation results (mIoU and total training time) of \textbf{CardioMix} on \textit{LUDB} under 1/4 labeled ratio (\textbf{MT} as base algorithm).
}
\setlength{\tabcolsep}{3pt}
\begin{tabular}{lcccc}
\toprule
\textbf{Method} & \textbf{Search} & \textbf{Conf. gating} & \textbf{mIoU (\%)} & \textbf{Train time (min)} \\
\midrule
\textbf{MT}& \xmark & \xmark & 78.7 & 14.7 \\
\midrule
\multirow{3}{*}{\textbf{+ L2U}} & Random & \xmark & 79.5 & 18.9 \\
& $\mathrm{Sim}(x_{a},x_{b})$ & \xmark & 79.6 & 20.2 \\
& $\mathrm{Sim}(y_{a},y_b)$ & \xmark & 80.0 & 19.5 \\
\midrule
\multirow{2}{*}{\textbf{+ U2L}}& $\mathrm{Sim}(y_{a},y_b)$ & \xmark & 80.6 & 20.4 \\
& $\mathrm{Sim}(y_{a},y_b)$ & \cmark & 81.3 & 20.4 \\
\midrule
+ \textbf{CardioMix} & $\mathrm{Sim}(y_{a},y_b)$ & \cmark & \textbf{82.2} & 20.1 \\
\bottomrule
\end{tabular}
\label{tab:directional_mix_ablation}
\end{table}

\subsection{Ablation Study}
\label{sec:ablation}

To validate the contributions of the key components in \textbf{CardioMix}, we conduct a series of ablation studies on the \textit{LUDB} dataset under a 1/4 labeled ratio using the \textbf{MT} as the base algorithm.
Each component is added incrementally to assess its individual contribution to both segmentation performance and computational overhead, as summarized in Table~\ref{tab:directional_mix_ablation}.

\subsubsection{Cardiac Pattern-Guided Search}
We first evaluate the importance of the cardiac pattern-guided segment search in L2U. We compare three selection criteria for the labeled key segment: (1) Random, which mimics vanilla CutMix but samples segments from the labeled set, (2) Signal similarity, which finds the segment with the highest cosine similarity in the signal space ($\mathrm{Sim}(x_a,x_b)$), and (3) Pattern similarity, which finds the segment with the highest IoU in segmentation space ($\mathrm{Sim}(y_a,y_b)$).
The results clearly show that selection based on the pattern similarity yields the best performance (mIoU=80.0\%), outperforming both random (79.5\%) and signal similarity-based (79.6\%) selection. This validates that matching in the segmentation space, which reflects cardiac patterns like \texttt{P-QRS-T} sequences, provides a more semantically aligned and effective data augmentation than matching in the raw signal space.

\subsubsection{U2L with Confidence Gating}
We then analyze the impact of U2L. Applying the cardiac pattern-guided U2L alone already improves performance (mIoU=80.6\%).
Confidence gating, which prevents noisy pseudo-labels from corrupting the labeled segment, is confirmed to be a critical component of U2L; integrating it further elevates performance to 81.3\%.
This gain underscores the necessity of selective and reliable information transfer from unlabeled data. 

\subsubsection{Full CardioMix Framework}
Combining all the components yields a full \textbf{CardioMix} framework.
\textbf{CardioMix} achieves the best result of 82.2\% with negligible computational overhead (e.g., approximately one minute of additional training time compared to L2U with Random).
This demonstrates the synergistic effect where cardiac pattern guidance, confidence gating, and bidirectional fusion jointly ensure physiologically valid, reliable, and comprehensive data utilization.

\subsubsection{Effect of Confidence Gating Threshold}
\begin{figure}[t]
    \centering
    \includegraphics[width=\columnwidth]{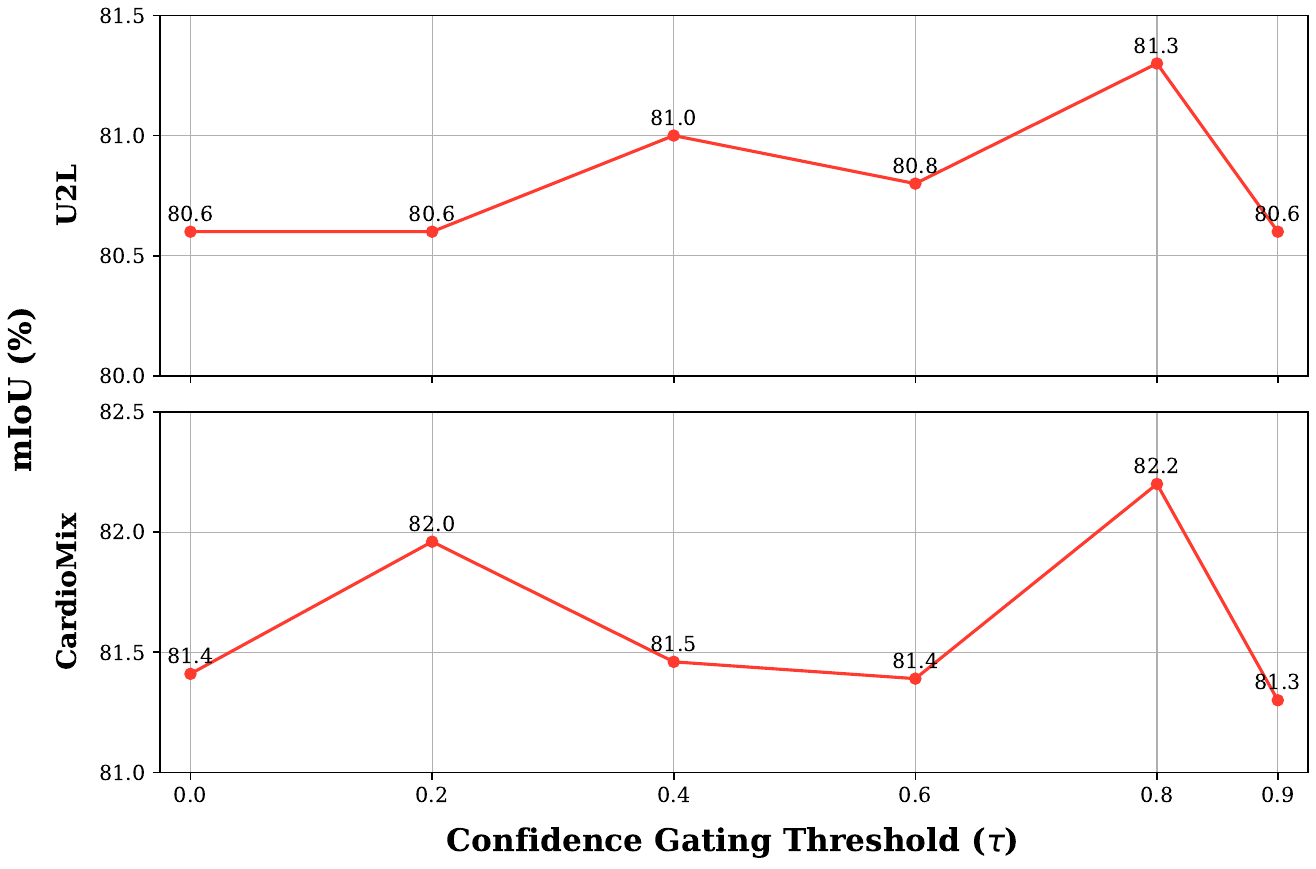}
    \caption{
        Sensitivity analysis on the confidence gating threshold $\tau$. The threshold gates the U2L fusion based on the average maximum softmax probability of the unlabeled segment. mIoU (\%) is reported for U2L only (top) and \textbf{CardioMix} (bottom) under varying values of $\tau$.
    }
    \label{fig:conf_thresh}
\end{figure}

The confidence gating threshold regulates the trade-off between the reliability and amount of information transfer during U2L.
To determine the optimal threshold and assess sensitivity, we perform a grid search over $\tau \in \{0.0, 0.2, 0.4, 0.6, 0.8, 0.9\}$ for both U2L only and \textbf{CardioMix}.
As illustrated in Fig.~\ref{fig:conf_thresh}, $\tau = 0.8$ achieves the best performance in both settings, outperforming the ungated case ($\tau = 0.0$) and reaffirming the role of selective gating.
Notably, increasing the threshold to 0.9 leads to a performance drop, likely because overly restrictive gating limits the contribution of informative unlabeled segments.

\subsection{Cardiac Pattern Consistency Analysis}
To assess the preservation of cardiac patterns, we measure the cardiac pattern consistency of augmented ECG signals after CutMix-based fusion, using the same experimental setting as the ablation study (\textit{LUDB}, 1/4 labeled ratio, \textbf{MT} as the base algorithm).

We identify pattern violations \emph{where a T-wave appears without a preceding QRS complex}, which are invalid in natural cardiac cycles. The consistency ratio is computed as the proportion of fused samples free from such violations.

\begin{figure}[t]
    \centering
    \includegraphics[width=\columnwidth]{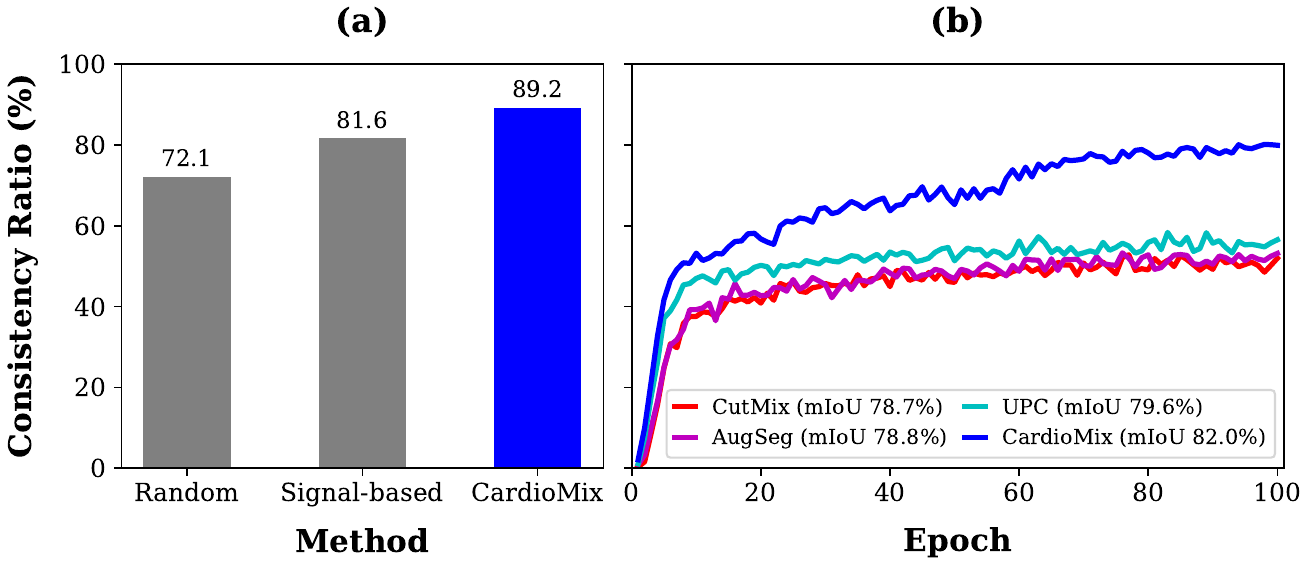} 
    \caption{
        Cardiac pattern consistency ratio of fused ECGs (\textit{LUDB}, 1/4 labeled ratio, \textbf{MT} as baseline algorithm). Consistency ratio measures the percentage of samples where all T-waves have preceding QRS complexes. (a) Consistency ratios after CutMix using ground-truth segmentation labels. (b) Consistency ratios after L2U during training with pseudo-labels.
    }
    \label{fig:quantitative_consistency}
\end{figure}

Fig.~\ref{fig:quantitative_consistency}(a) shows consistency ratios when applying CutMix with ground-truth segmentation labels, with Random, Signal-based, and \textbf{CardioMix} corresponding to three selection strategies evaluated in Section~\ref{sec:ablation}. The consistency ratio ranking closely mirrors the performance ranking in Table~\ref{tab:directional_mix_ablation}: \textbf{CardioMix}, which utilizes pattern similarity-based selection, achieved the highest consistency ratio (89.2\%), followed by signal similarity-based (81.6\%) and random selection (72.1\%). 

\begin{figure}[t]
    \centering
    \includegraphics[width=\columnwidth]{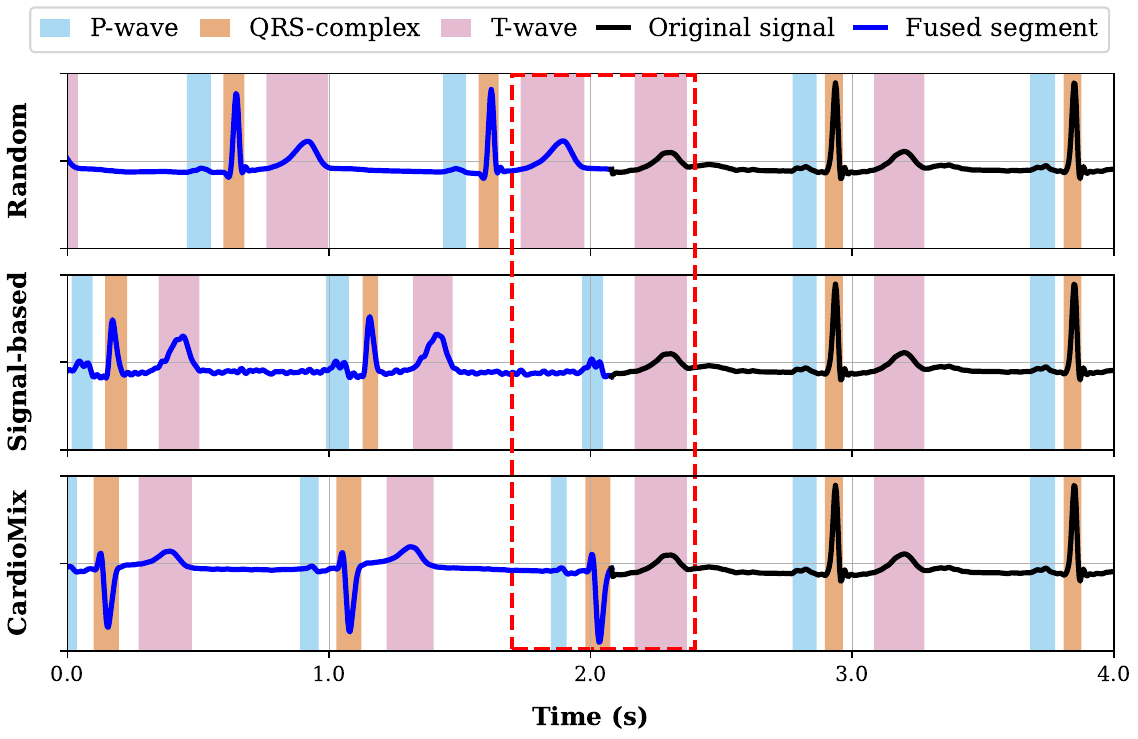}
    \caption{
        Qualitative example 1. Random selection produces a \texttt{T-T} sequence and signal-based selection produces a \texttt{P-T} sequence within the fused region (red dashed box), while \textbf{CardioMix} preserves the natural \texttt{P-QRS-T} cycle.
    }
    \label{fig:qualitative_ex1}
\end{figure}

\begin{figure}[t]
    \centering
    \includegraphics[width=\columnwidth]{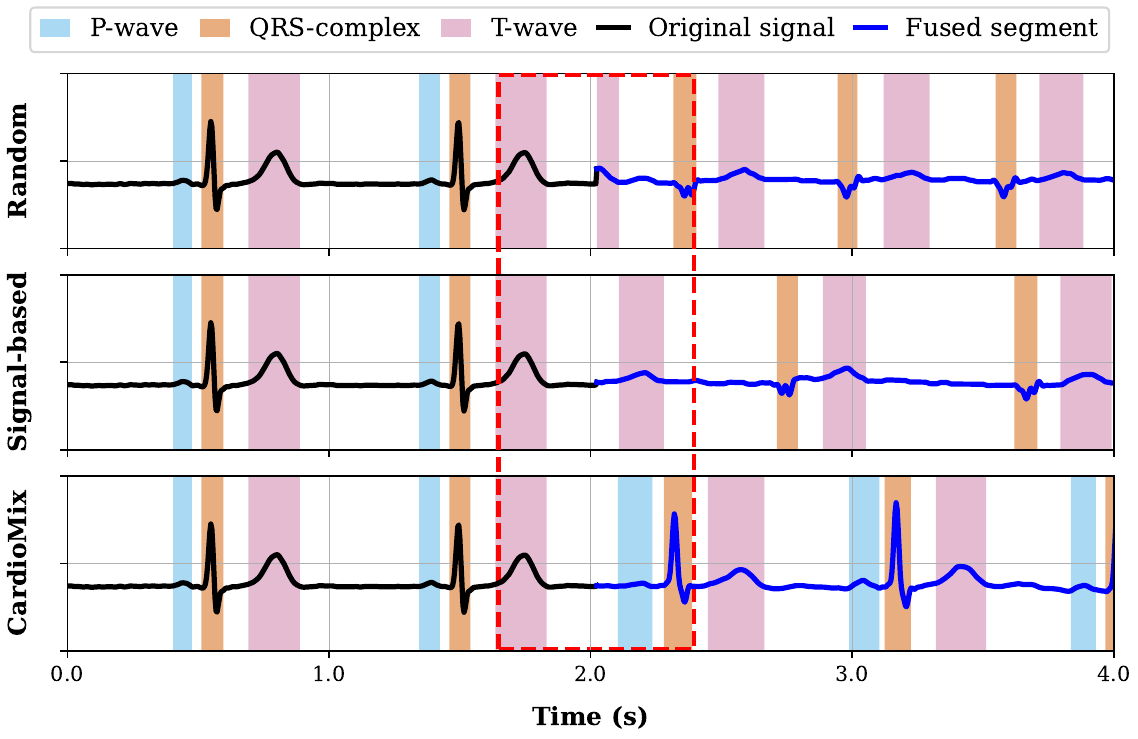}
    \caption{
        Qualitative example 2. Both random and signal-based selection produce \texttt{T-T} sequences within the fused region (red dashed box), while \textbf{CardioMix} maintains valid cardiac sequences.
    }
    \label{fig:qualitative_ex2}
\end{figure}

Fig.~\ref{fig:qualitative_ex1} and Fig.~\ref{fig:qualitative_ex2} present qualitative examples comparing the three selection strategies.
Random selection is inherently prone to producing physiologically implausible patterns, such as \texttt{T-T} sequences, by carelessly placing T-wave segments consecutively. Signal-based selection, which matches low-level waveform appearances through cosine similarity, also fails to account for the underlying cardiac phase, potentially leading to irregular sequences like \texttt{P-T} patterns. In contrast, \textbf{CardioMix} consistently preserves the valid \texttt{P-QRS-T} structure by matching segments based on their composition in the segmentation space. This approach ensures that fused segments remain semantically and physiologically coherent, effectively supporting more robust consistency regularization.

Fig.~\ref{fig:quantitative_consistency}(b) tracks the consistency ratio after L2U during training. All methods show increasing consistency ratio from zero, reflecting gradual improvement of pseudo-label quality. \textbf{CardioMix} maintains the highest consistency ratio throughout training, followed by \textbf{UPC}, \textbf{AugSeg}, and \textbf{CutMix}, consistent with their performance rankings.

These results suggest that preserving cardiac patterns during fusion correlates with improved segmentation performance. 

\section{Discussion}

\subsection{Overall Findings}
\textbf{CardioMix} integrates cardiac pattern guidance into a bidirectional fusion framework, ensuring that augmented samples maintain physiologically valid \texttt{P-QRS-T} sequences throughout the fusion process, simultaneously stabilizing pseudo-label quality through L2U and enriching the limited labeled set with realistic variations through U2L.
The cardiac pattern consistency analysis supports this design, suggesting a notable correlation between pattern preservation and segmentation performance.

These properties translate into consistent performance gains in the ablation study, suggesting that the effectiveness of \textbf{CardioMix} stems not from any single component, but from the synergy between cardiac pattern guidance and bidirectional fusion. This synergy further holds across diverse datasets, labeled ratios, and base algorithms, surpassing methods that rely on random segment selection or unidirectional fusion.

An exception is observed on the \textit{Zhejiang} dataset, where \textbf{UPC} occasionally outperforms \textbf{CardioMix}. We attribute this to the prevalence of zero-padded regions disrupting the cardiac pattern composition and limiting the effectiveness of pattern-guided segment matching.

Together, these results demonstrate that physiological structure preservation serves as a critical factor for effective semi-supervised learning on structured biomedical time series such as ECG.

\subsection{Clinical Implications}
Accurate ECG delineation is a prerequisite for the automated computation of clinically critical intervals such as the PR interval, QRS duration, and QT interval, which are central to the diagnosis of cardiovascular conditions, including arrhythmias and conduction disorders. The MAE results indicate that \textbf{CardioMix} yields more precise interval estimations beyond improvements in mIoU. The demonstrated data efficiency, achieving competitive performance with half the labeled data, has direct practical value, as obtaining high-quality ECG annotations requires significant clinical expertise and is inherently costly.

\subsection{Limitations and Future Work}
This study has several limitations.
First, \textbf{CardioMix} introduces a dependency on pseudo-label quality, as noisy early-stage predictions may affect the reliability of the cardiac pattern-guided search and confidence gating. While our results suggest that these mechanisms jointly mitigate this issue in practice, further investigation into robust training strategies under severe pseudo-label noise remains an open direction.
Second, the evaluation datasets in this study primarily consist of clinical ECGs recorded under controlled conditions. Whether \textbf{CardioMix} generalizes to ECGs from patients with diverse arrhythmias or structural cardiac diseases, or to signals acquired from mobile and wearable devices with higher noise levels, remains to be validated.
Third, the cardiac pattern-guided search is tailored to the \texttt{P-QRS-T} structure of ECG signals. Although the underlying principle of leveraging domain-specific structural patterns is broadly applicable, direct extension to other time-series domains will require adaptation to their respective structural characteristics.

Building on these limitations, future work will prioritize validating \textbf{CardioMix} on more clinically diverse ECG datasets, including those from patients with arrhythmias and structural cardiac diseases, as well as signals from mobile and wearable devices. We also plan to explore combining \textbf{CardioMix} with self-supervised pretraining, as leveraging large-scale unlabeled data through self-supervised representations could further enhance performance under label-scarce conditions. Finally, extending the principle of pattern-guided fusion to other structured time-series domains such as EEG, PPG, and respiratory waveforms represents a promising direction for future work.

\section{Conclusion}
We introduce \textbf{CardioMix}, a semi-supervised ECG segmentation framework that leverages cardiac pattern guidance to ensure physiologically meaningful data fusion. By bidirectionally exchanging information between labeled and unlabeled sets, \textbf{CardioMix} consistently outperforms existing CutMix-based fusion strategies across diverse datasets, labeled ratios, and base algorithms, while demonstrating remarkable data efficiency. To the best of our knowledge, this study also presents the first comprehensive benchmark evaluating CutMix-based fusion strategies within semi-supervised ECG delineation. We hope \textbf{CardioMix} and its underlying principle of pattern-preserving fusion inspire future work in semi-supervised learning for structured biomedical signals.

\section{Data Availability}
All datasets used in this study are publicly available. We utilized the preprocessed versions provided by the \textsf{SemiSegECG} benchmark, which is accessible at \url{https://github.com/vuno/semi-seg-ecg}.

\section*{References}

\bibliographystyle{IEEEtran}
\bibliography{cardiomix}

\newpage

\renewcommand{\thetable}{S\arabic{table}}
\renewcommand{\thefigure}{S\arabic{figure}}
\setcounter{table}{0}
\setcounter{figure}{0}

\appendices
\section{Detailed Performance Analysis}
\label{appendix-ext_rst}

We further present extensive benchmarking results on \textit{QTDB} (Table~\ref{tab:bench-qtdb}), \textit{ISP} (Table~\ref{tab:bench-isp}), and \textit{Zhejiang} (Table~\ref{tab:bench-zhejiang}). \textbf{CardioMix} consistently achieves strong performance on \textit{QTDB} and \textit{ISP} across various labeled ratios and SemiSeg algorithms. An exception is observed on the \textit{Zhejiang} dataset, where \textbf{UPC} occasionally exhibits a marginal advantage, as discussed in Section~\ref{sec:sota} of the main paper.

\begin{table}[h]
\centering
\caption{
    \textit{QTDB} benchmarking results (mIoU, \%) of CutMix-based fusion strategies across all base algorithms.
}
\begin{tabular}{l|cccc}
\toprule
\multirow{2}{*}{\textbf{Methods}} & 1/16 & 1/8 & 1/4 & 1/2 \\
& (\textit{N}=22)& (\textit{N}=56)& (\textit{N}=114) & (\textit{N}=222) \\
\midrule
\textbf{Scratch}& 39.1±0.4& 51.0±1.7& 61.9±0.3& 68.0±0.8
\\
\midrule
\textbf{MT} & 55.1±0.2& 56.9±1.0& 64.9±0.2& 69.4±0.2
\\
+ \textbf{CutMix} & 55.1±0.4& 57.8±1.3& 66.1±0.9& 69.2±0.4
\\
+ \textbf{AugSeg} & 55.9±1.0& 57.3±1.0& 66.5±0.4& 69.9±0.9
\\
+ \textbf{UPC} & 56.9±0.8& 58.7±0.9& 66.0±0.7& 70.0±0.4
\\
+ \textbf{CardioMix} & \textbf{60.2±0.6}& \textbf{62.8±0.4}& \textbf{68.4±0.5}& \textbf{71.3±0.3}\\
\midrule
\textbf{FixMatch}& 48.4±1.6& 53.6±0.6& 63.8±0.5& 69.1±0.4
\\
+ \textbf{CutMix} & 51.8±1.2& 55.7±1.2& 64.4±0.5& 68.9±0.5
\\
+ \textbf{AugSeg} & 50.6±0.9& 54.9±1.2& 64.9±0.9& 69.5±0.5
\\
+ \textbf{UPC} & 52.4±0.9& 57.0±0.2& 64.9±0.1& 70.1±1.2
\\
+ \textbf{CardioMix} & \textbf{52.9±0.8}& \textbf{62.0±0.9}& \textbf{67.6±0.8}& \textbf{72.0±0.5}\\
\midrule
\textbf{CPS}& 45.8±4.5& 56.7±1.2& 60.1±1.3& 68.0±0.7
\\
+ \textbf{CutMix} & \textbf{56.4±1.6}& 57.2±0.6& 65.9±0.5& 70.0±0.6
\\
+ \textbf{AugSeg} & 55.0±2.3& 57.4±0.9& \textbf{66.4±0.8}& 69.8±0.6
\\
+ \textbf{UPC} & 48.3±1.5& 56.9±0.8& 64.3±1.3& 69.8±0.5
\\
+ \textbf{CardioMix} & 51.0±2.2& \textbf{59.2±1.4}& 65.7±0.8& \textbf{70.9±0.7}\\
\midrule
\textbf{ReCo} & 45.6±0.5& 49.7±2.3& 58.6±0.9& 61.7±1.3
\\
+ \textbf{CutMix} & 48.2±0.6& 53.2±0.8& 60.3±1.2& 63.1±0.3
\\
+ \textbf{AugSeg} & 51.2±1.3& 55.2±0.7& 62.7±0.4& 66.3±0.5
\\
+ \textbf{UPC} & \textbf{52.9±1.7}& 57.1±0.6& 64.0±1.0& 67.9±0.5
\\
+ \textbf{CardioMix} & 52.7±0.6& \textbf{60.4±0.7}& \textbf{65.0±0.4}& \textbf{68.4±0.4}\\
\midrule
\textbf{ST++} &50.6±1.3& 55.2±0.9& 64.1±1.4& 69.2±0.5
\\
+ \textbf{CutMix} & 51.2±0.7& 55.6±1.2& 65.4±1.0& 69.8±0.1
\\
+ \textbf{AugSeg} & 52.8±2.4& 56.5±1.0& 65.4±0.7& 70.5±0.4
\\
+ \textbf{UPC} & 51.9±1.0& 57.7±0.5& \textbf{66.2±1.0}& 70.2±0.1
\\
+ \textbf{CardioMix} & \textbf{52.9±2.0}& \textbf{61.5±0.2}& \textbf{66.2±0.3}& \textbf{71.9±0.6}\\
\bottomrule
\end{tabular}
\label{tab:bench-qtdb}
\end{table}

\begin{table}[t]
\centering
\caption{
    \textit{ISP} benchmarking results (mIoU, \%) of CutMix-based fusion strategies across all base algorithms.
}
\begin{tabular}{l|cccc}
\toprule
\multirow{2}{*}{\textbf{Methods}} & 1/16 & 1/8 & 1/4 & 1/2 \\
& (\textit{N}=228) & (\textit{N}=468) & (\textit{N}=948) & (\textit{N}=1896) \\
\midrule
\textbf{Scratch}& 69.7±0.6& 73.0±0.2& 76.6±0.2& 78.3±0.2
\\
\midrule
\textbf{MT} & 74.6±0.1& 75.7±0.3& 78.1±0.3& 78.8±0.3
\\
+ \textbf{CutMix} & 72.9±0.4& 75.0±0.3& 78.4±0.4& 80.1±0.2
\\
+ \textbf{AugSeg} & 73.3±0.6& 75.2±0.7& 78.3±0.4& 80.1±0.5
\\
+ \textbf{UPC} & 73.8±0.1& 76.2±0.3& 79.6±0.7& 81.6±0.4
\\
+ \textbf{CardioMix} & \textbf{75.0±0.7}& \textbf{78.9±0.3}& \textbf{81.4±0.2}& \textbf{82.5±0.1}\\
\midrule
\textbf{FixMatch}& \textbf{74.7±0.1}& 75.7±0.2& 78.0±0.2& 79.2±0.3
\\
+ \textbf{CutMix} & 73.5±0.5& 75.0±0.4& 78.0±0.3& 79.0±0.3
\\
+ \textbf{AugSeg} & 73.6±0.2& 75.2±0.6& 78.1±0.3& 79.6±0.0
\\
+ \textbf{UPC} & 73.2±0.5& 74.8±0.3& 78.9±0.1& 80.5±0.6
\\
+ \textbf{CardioMix} & 73.9±0.5& \textbf{76.4±0.3}& \textbf{81.1±0.2}& \textbf{82.4±0.1}\\
\midrule
\textbf{CPS}& 72.7±0.2& 74.2±0.4& 78.5±0.1& 79.5±0.3
\\
+ \textbf{CutMix} & \textbf{74.6±0.4}& 75.3±0.4& 78.6±0.2& 80.8±0.2
\\
+ \textbf{AugSeg} & 74.4±0.3& 75.4±0.4& 78.5±0.3& 80.9±0.4
\\
+ \textbf{UPC} & 73.9±0.1& \textbf{76.2±0.5}& \textbf{80.9±0.1}& 82.1±0.3
\\
+ \textbf{CardioMix} & 74.2±0.6& 75.4±0.1& 80.7±0.7& \textbf{82.2±0.2}\\
\midrule
\textbf{ReCo} & 71.5±0.2& 73.3±0.2& 76.1±0.2& 77.0±0.3
\\
+ \textbf{CutMix} & 73.5±0.1& 74.8±0.1& 77.2±0.2& 78.0±0.2
\\
+ \textbf{AugSeg} & 73.3±0.4& 74.7±0.2& 77.4±0.2& 78.1±0.2
\\
+ \textbf{UPC} & 73.8±0.1& \textbf{75.3±0.2}& \textbf{77.8±0.3}& \textbf{78.7±0.1}\\
+ \textbf{CardioMix} & \textbf{73.9±0.1}& 75.2±0.3& \textbf{77.8±0.1}& 78.3±0.2
\\
\midrule
\textbf{ST++} &73.6±0.7& 75.3±0.4& 78.2±0.1& 79.5±0.1
\\
+ \textbf{CutMix} & 73.8±0.1& 75.7±0.3& 78.3±0.4& 80.4±0.1
\\
+ \textbf{AugSeg} & 73.9±0.2& 75.5±0.2& 78.8±0.4& 80.5±0.2
\\
+ \textbf{UPC} & 73.8±0.6& 75.7±0.2& 79.9±0.2& \textbf{81.7±0.1}\\
+ \textbf{CardioMix} & \textbf{74.2±0.1}& \textbf{76.7±0.4}& \textbf{80.1±0.2}& 81.6±0.5\\
\bottomrule
\end{tabular}
\label{tab:bench-isp}
\end{table}

\begin{table}[t]
\centering
\caption{
    \textit{Zhejiang} benchmarking results (mIoU, \%) of CutMix-based fusion strategies across all base algorithms.
}
\begin{tabular}{l|cccc}
\toprule
\multirow{2}{*}{\textbf{Methods}} & 1/16 & 1/8 & 1/4 & 1/2 \\
& (\textit{N}=144) & (\textit{N}=300) & (\textit{N}=600) & (\textit{N}=1200) \\
\midrule
\textbf{Scratch}& 68.9±0.9& 73.6±0.5& 76.1±0.2& 79.9±0.6
\\
\midrule
\textbf{MT} & 78.8±0.8& 81.1±0.3& 81.8±0.4& 83.5±0.1
\\
+ \textbf{CutMix} & 78.9±0.6& 81.0±0.0& 82.2±0.1& 83.6±0.3
\\
+ \textbf{AugSeg} & 79.4±0.5& 81.5±0.3& 82.6±0.4& 83.6±0.2
\\
+ \textbf{UPC} & 79.8±0.1& \textbf{82.1±0.2}& \textbf{82.9±0.3}& \textbf{83.8±0.3}
\\
+ \textbf{CardioMix} & \textbf{81.1±0.1}& 82.0±0.0& \textbf{82.9±0.1}& 83.7±0.0
\\
\midrule
\textbf{FixMatch}& 75.5±0.5& 80.1±0.1& 81.7±0.3& 83.2±0.3
\\
+ \textbf{CutMix} & 77.5±0.3& 80.9±0.1& 81.9±0.1& 83.0±0.1
\\
+ \textbf{AugSeg} & 77.5±0.7& 80.7±0.4& 81.7±0.1& 83.2±0.1
\\
+ \textbf{UPC} & 80.0±0.2& \textbf{82.0±0.2}& \textbf{82.7±0.2}& \textbf{83.8±0.3}
\\
+ \textbf{CardioMix} & \textbf{80.2±0.1}& 81.5±0.1& 82.4±0.2& 83.3±0.1
\\
\midrule
\textbf{CPS}& 73.8±0.6& 79.8±0.5& 81.7±0.2& 82.7±0.1
\\
+ \textbf{CutMix} & 78.9±0.6& 81.4±0.3& 82.1±0.1& 83.3±0.2
\\
+ \textbf{AugSeg} & 78.1±0.6& 81.1±0.1& 82.1±0.3& 83.3±0.1
\\
+ \textbf{UPC} & \textbf{80.7±0.2}& \textbf{82.7±0.1}& \textbf{82.9±0.3}& \textbf{83.8±0.1}
\\
+ \textbf{CardioMix} & 80.2±0.1& 82.1±0.1& 82.6±0.2& 83.4±0.1
\\
\midrule
\textbf{ReCo} & 67.9±0.8& 68.3±0.1& 66.7±4.2& 66.4±2.9
\\
+ \textbf{CutMix} & 68.9±2.8& 69.6±2.8& 70.5±0.7& 67.8±3.4
\\
+ \textbf{AugSeg} & 71.9±0.3& 72.6±0.3& 72.6±0.4& 73.0±0.3
\\
+ \textbf{UPC} & \textbf{74.3±0.2}& \textbf{76.0±0.1}& \textbf{76.4±0.1}& \textbf{76.9±0.1}
\\
+ \textbf{CardioMix} & 73.6±0.1& 74.7±0.1& 74.8±0.3& 74.8±0.2
\\
\midrule
\textbf{ST++} &74.5±0.5& 78.6±0.4& 80.7±0.6& 83.3±0.2
\\
+ \textbf{CutMix} & 76.4±0.5& 80.3±0.7& 81.7±0.1& 83.7±0.3
\\
+ \textbf{AugSeg} & 76.5±0.5& 80.4±0.8& 81.9±0.4& 83.7±0.3
\\
+ \textbf{UPC} & \textbf{79.8±0.6}& \textbf{81.3±0.3}& \textbf{83.0±0.2}& \textbf{84.0±0.1}
\\
+ \textbf{CardioMix} & 77.9±0.3& 80.5±0.1& 82.3±0.3& 83.5±0.4
\\
\bottomrule
\end{tabular}
\label{tab:bench-zhejiang}
\end{table}

\end{document}